\documentclass{melba}

\usepackage{mwe} 

\usepackage{amsmath,amsfonts}

\melbaid{YYYY:NNN}  
\doi{10.59275/j.melba.2024-AAAA}
\melbaauthors{Michele Bufano and Elmar Kotter}  
\email{michele.bufano@uniklinik-freiburg.de}
\volume{2}
\firstpageno{1}  
\melbayear{2025}  
\datesubmitted{yyyy-m1-d1}  
\datepublished{yyyy-m2-d2}  

\ShortHeadings{Deep classification algorithm for De-identification of DICOM medical images}{M.Bufano and E.Kotter}

\title{Deep classification algorithm for De-identification of DICOM medical images}

\author{
	\firstname Michele \surname Bufano\aff{1}\orcid{0009-0000-5067-9814},
	\name Elmar Kotter\aff{1}\orcid{0000-0001-9022-6000}
}
\affiliations{
	\num 1 \addr University Medical Center Freiburg, Freiburg, DE
}

\abstract{
	\paragraph{Background}: De-identification of DICOM (Digital Imaging and Communi-cations in Medicine) files is an essential component of medical image research. Personal Identifiable Information (PII) and/or Personal Health Identifying Infor-mation (PHI) need to be hidden or removed due to legal reasons. According to the Health Insurance Portability and Accountability Act (HIPAA) and privacy rules, also full-face photographic images and any compa-rable images are direct identifiers and are considered protected health information that also need to be de-identified.
	\paragraph{Objective}: The study aimed to implement a method that permit to de-identify the PII and PHI information present in the header and burned on the pixel data of DICOM.
	\paragraph{Methods}: To execute the de-identification, we implemented an algorithm based on the  \emph{safe harbor method}, defined by HIPAA. Our algorithm uses input customizable parameter to classify and then possibly de-identify individual DICOM tags.
	\paragraph{Results}: The most sensible information, like names, history, personal data and institution were successfully recognized.
	\paragraph{Conclusions}: We developed a python algorithm that is able to classify infor-mation present in a DICOM file. The flexibility provided by the use of customi-zable input parameters, which allow the user to customize the entire process de-pending on the case (e.g., the language), makes the entire program very promis-ing for both everyday use and research purposes.
\\
Our code is available at~\url{https://github.com/rtdicomexplorer/deep_deidentification}.}


\begin{document}

\twocolumn[\maketitle]

\section{Introduction}
	\enluminure{R}{esearch} in the health sector needs data to enable experimentation and creation of new and better tools to improve patient care. For this reason, sharing data is important. How-ever, sharing personal information can only occur with the patient's consent (HIPAA)
	DICOM Standard  provides in the Application Level Confidentiality Profile and Options Table E.1-1 a baseline to perform an anonymization process. However, in this way, many data necessary for research are lost.

\section{De-identification}
De-identification ensures the privacy of the patient's personal data, but at the same time allows the sharing of information necessary for research. The de-identification process involves the removal of patient personal information (PII) and patient health information (PHI), whether it is present in the DICOM header or burned on the pixel data. It is regulated by HIPAA, which provides two methods (see Figure 1)
\\
Our algorithm implements the safe harbor method. Starting from a list of key-words to search, it performs an in-depth search through the DICOM tags  present in the header, classifying each of them. In a subsequent phase, the de-identification is performed by actions, predefined or customized, based on the type of classification.
\\

\section{Methods}
Our approach involves an initialization phase and three subsequent steps which concern:
\begin{itemize}
\item{Classification of patient collection}
\item {De-identification of single DICOM file header}
\item {Removing burned sensible information present on the pixel data for the single DICOM file}
\end{itemize}
	\subsection{Sorting DICOM data}
		All DICOM files were sorted in a collection according the DICOM hierarchy, which is organized in four levels: patient, study, series and instance (see Fig.2).
	\subsection{Initialization}
	For its execution, our algorithm needs a list of \emph{searching-parameters} that are used to classify the DICOM tags, a list of sensible DICOM tags which contain sensible informations, a list of actions to perform depending on the previous classification and a list of default values to use for the de-identification.
	\subsubsection{Searching-parameter}
	To define them we implemented an algorithm (deep-search-algorithm) that recovered the words that preceded the names of people, institutions, places and telephone numbers. We applied this algorithm on the 1693 files of the training data. In this way, we defined 3 groups of input words data to search for the matching (see Tables 1,2,3)
	\subsubsection{Sensible DICOM tags}
	The user can define a list of DICOM tags to de-identify (custom DICOM tags), whose values could be present elsewhere even burned on the pixel data.
	\subsubsection{Actions}
	We defined 2 types:
	\begin{itemize}
		\item{Default actions, defined in accordance with the Table E.1 present in DICOM standard}
		\item {Custom actions, which override the default actions. (See Table 5)}
	\end{itemize}
	
	\subsubsection{Default values}
	According with the value representation (VR) of the DICOM tags, which must be used to replace the values to be de-identified. (See Table 6)
	\subsection{De-identification}
	The program starts scanning all the files and proceeds with the classification of each DICOM tag through a \emph{classification-step}, and then the relative de-identification step executed (See Figure. 3)
	\\
	Task of the classification action is also to recover the values of the DICOM tags present sensible tag list given. (See Figure 4)
	\subsection{Removing Burned information}
	In this step, we decided use \emph{keras-ocr} algorithm, a ML-Package that permits to recognize all text burned on a picture.
	The algorithm recognizes and retrieves the boundary box of each text.
	Using of a similarity text package, \emph{thefuzz} , all burned text recognized were compared with the sensible information saved during the classification-action in the previous step.
	\\
	The matching was considered positive for a similarity score  \textgreater 49\%, in this case the text on the pixel data was deleted by overwriting the pixels inside the boundary box calculated by the model. The color used to fill the boundary box is black. To calcu-late the relative HU (Hounsfield Value ) we used the tags information defined in the section C.7.6.3 of DICOM standard.
	Since the keras-ocr algorithm works on 8-bit per channel images, we scaled the calculated DICOM pixel data to 8 bits using also the tags information defined in the section C.7.6.3 of the DICOM standard.
\section{Results}
	During the validation phase, where have been de-identified 23961 files containing 216 patients, we updated the search parameters list found previously with the \emph{deep-search-algorithm}. We also added the less recurring words; this allowed the algorithm to improve the score from 98.45 to 99.86.
	The results were confirmed in the final test phase, where have been de-identified 322 patients for 29660 files.
	The use of texts to classify individual DICOM tags worked very well for recognizing personal and institutional data information. The results regarding geographical data were slightly worse.
	\subsection{Other languages than English}
	We also used the algorithm to de-identify files with different languages like German and Italian. The results were confirmed. The deep-search-algorithm found different texts than the English ones.
	The words were the translation of the English words. For this reason, the end user could also directly enter the words to search for.
	\subsection{Burned texts}
	Regarding the recognition and removal of burned information present on the pixel data, here too the results were excellent for DX, CR and MG modalities with removal of all personal patient information. (See Fig.6)
	For other modalities (e.g., CT) the keras-ocr algorithms did not work very well because some pixel intensities variations were misinterpreted as text.
\section{Conclusions}
	The de-identification was successful for almost all PII and PHI information. Improvements were later achieved by inserting words present in the private tags such as manufacturer name. 
	\\
	The recognition and the following removal of texts burned on the pixel data de-pends on the efficiency of the OCR algorithm used. Even in this case, the perfor-mance would surely improve with targeted fine-tuning operations using this type of images.
	\\
	The algorithm is very flexible, it can be adapted by the end user in any situation, not only in de-identification, but also in anonymization processes simply by varying the input parameters. The separation of the entire procedure into different steps allows improving the in-dividual steps independently of each other. All this makes the program very promis-ing.
	\\
	The procedures that constitute the algorithm are very simple, and independent from the development environment, therefore easily implementable in other comput-er-languages, where it is possible to use comfortable GUI (Graphic User Interface), which allow easy use by medical personnel without requiring particular computer knowledge. A version is currently being implemented in MAUI .
	\\
	The algorithm was developed in python. The code is open source  and is posted on GitHub. To parse the DICOM files, we used \emph{pydicom} (a python package that allows managing DICOM files). All input parameters are defined in a json  file.
	\\
	As far is the hardware concerned, there are no special requirements. The de-identification of the entire set of test data took 95 minutes using a WIN10 22H2 64bit with an Intel I711th Gen 2.5 GHz and 64GB RAM.

\section{Table and Figures}
	\begin{table}[ht] 
		\centering
				\begin{tabular}{lcr}
		\\
			\text{Institution text}  
			\\
			\hline
			clinic, hospital, department, 
			\\medical, uiversity, clinician,
			\\hospice, memorial, follow up
		\end{tabular}
		\caption{List of words commonly used in medical reports to refer to persons or institutions}

	\end{table}

	\begin{table}[ht] 
		\centering
		\begin{tabular}{lcr}
		\\
			\text{Geographic text}  
			\\
			\hline
			street, road, route, avenue, 
			\\straße, allee, via, corso
		\end{tabular}
		\caption{List of words commonly used to refer to an address or a place}

	\end{table}

	\begin{table}[ht] 
		\centering
		\begin{tabular}{lcr}
		\\
			\text{Preposition text}  
			\\
			\hline
			for, to, on, call, at, by, prof, dr
		\end{tabular}
		\caption{List of prepositions commonly used for list of prepositions normally used before names, address-es and places, telephone numbers}

	\end{table}
	
	\begin{table}[ht] 
		\centering
		\begin{tabular}{lcr}
		\\
			\text{Sensible Tag List}  
			\\		
			\hline
			\textbf{DICOM TAG} & \textbf{Name}
			\\0008,0020 & Study date
			\\0008,0021 & Series date
			\\0008,0090 & Referring physician
			\\0008,1048 & Physician record
			\\0008,1050 & Performing physician
			\\0008,1070 & Operator's name
			\\0010,0010 & Patient's name
			\\0010,0020 & Patient ID
			\\0010,0030 & Physician Patient's birth date
			\\0040,0075 & Verifying observer

		\end{tabular}
		\caption{List of DICOM tags, which contain sensible data to search and remove also in other tags or burned on the pixel data}

	\end{table}

	\begin{table}[ht] 
		\centering
		\begin{tabular}{lcr}
		\\
		\text{Custom actions}  
			\\
			\hline
			\\Shift date by value for Date Tags
			\\Shift time by value for Time Tags
			\\Replace value for UI Tags 
			\\Keep value for Patient's Age
		\end{tabular}
		\caption{List of custom actions used to manage DICOM tags}

	\end{table}

	\begin{table}[ht] 
		\centering
		\begin{tabular}{lcr}
		\\
		\text{Default values}  
			\\
			\hline
			\textbf{VR} & \textbf{Value}
			\\DT & 00010101010101
			\\TM & 000000.000000
			\\DA & 00010101
			\\LO, LT, SH, PN, CS, ST, UT, UN & Anonymized
			\\FD, FL, SS, US, SL, UL, DS, IS & 0
		\end{tabular}
		\caption{List of custom actions used to manage DICOM tags}

	\end{table}

	\begin{figure}[ht]
		\centering
		\includegraphics[width=0.5\linewidth]{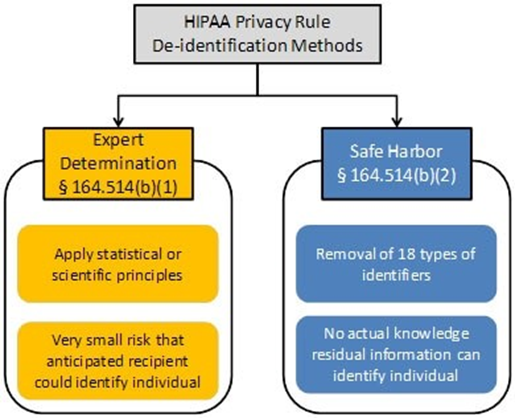}
		\caption{Two methods to achieve de-identification in accordance with the HIPAA Privacy Rule (source HIPAA)}
	\end{figure}
	\begin{figure}[ht]
		\centering
		\includegraphics[width=0.5\linewidth]{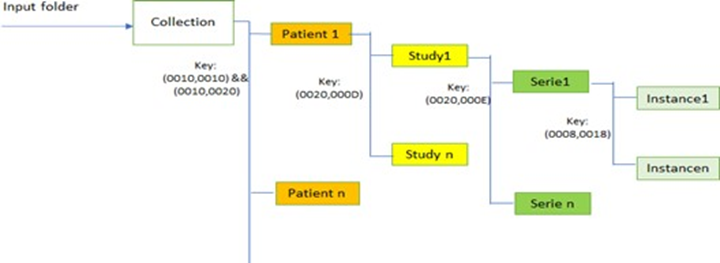}
		\caption{DICOM hierarchy}
	\end{figure}

	\begin{figure}[ht]
		\centering
		\includegraphics[width=0.5\linewidth]{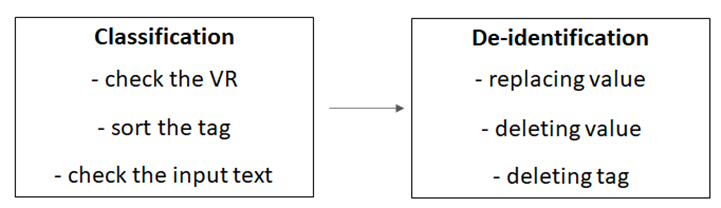}
		\caption{Header de-identification schema}
	\end{figure}

	\begin{figure}[ht]
		\centering
		\includegraphics[width=0.5\linewidth]{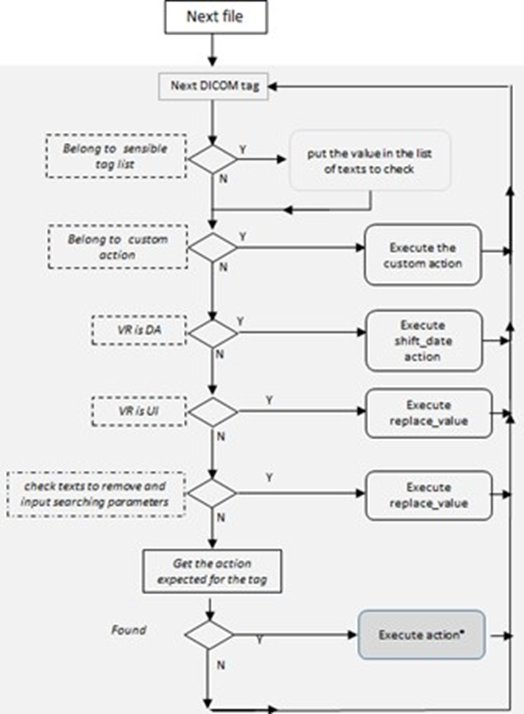}
		\caption{Workflow header de-identification}
	\end{figure}

	\begin{figure}[ht]
		\centering
		\includegraphics[width=0.5\linewidth]{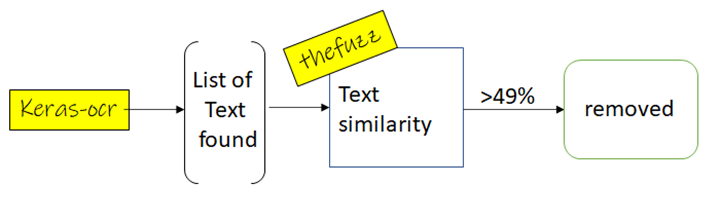}
		\caption{Burned texts classification}
	\end{figure}

	\begin{figure}[ht]
		\centering
		\includegraphics[width=0.5\linewidth]{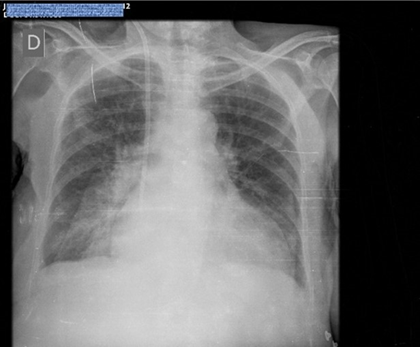}
		\caption{Image with sensible information burned}
	\end{figure}

	\begin{figure}[ht]
		\centering
		\includegraphics[width=0.5\linewidth]{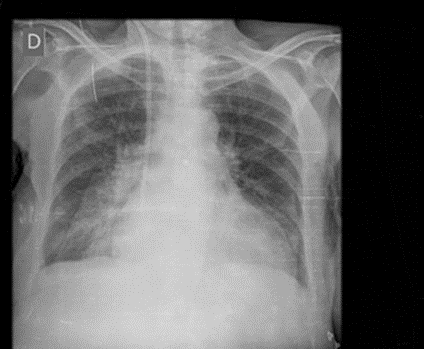}
		\caption{Image after the removing burned information}
	\end{figure}


%
\ethics{The work follows appropriate ethical standards in conducting research and writing the manuscript, following all applicable laws and regulations regarding treatment of animals or human subjects.}

\coi{We declare we don't have conflicts of interest.}

\data{Authors submitting articles to \textsc{Melba} are required to include a Data Availability Statement in their manuscripts. The Data Availability Statement should clearly indicate whether the data supporting the findings of the study are available and, if so, how readers can access them. If the data are not available, authors should provide a brief justification for not sharing the data.}

\bibliography{Anonymization of DICOM electronic medical records for radiation therapy. Computers in Biology and Medicine 53(2014)134–140
from Wayne Newhauser et al.}


\clearpage

\end{document}